\documentclass[conference, letterpaper]{IEEEtran}

\ifCLASSINFOpdf
   \usepackage[pdftex]{graphicx}
\else
\fi

\usepackage[cmex10]{amsmath}
\usepackage{color}
\usepackage{amsmath,amssymb,amsfonts}
\usepackage{algorithmic}
\usepackage{graphicx}
\usepackage{textcomp}
\usepackage{xcolor}
\usepackage{fontawesome}
\usepackage{url}
\usepackage{fancyhdr}
\usepackage[caption=false,font=footnotesize]{subfig}

\renewcommand{\thispagestyle}[2]{}

\fancypagestyle{plain}{
        \fancyhead{}
        \fancyhead[C]{first page center header}
        \fancyfoot{}
        \fancyfoot[C]{first page center footer}
}
\begin{document}

%
\title{QUANTUM MACHINE LEARNING APPLIED TO THE CLASSIFICATION OF DIABETES\\}

\author{\IEEEauthorblockN{Hancco-Quispe Juan Kenyhy }
\IEEEauthorblockA{Faculty of Statistic and Computer Engineering,\\ Universidad Nacional del Altiplano de Puno, P.O. Box 291\\ Puno - Peru.\\
Email: jkenyhyhq@gmail.com }
\and
\IEEEauthorblockN{Borda-Colque Jordan Piero }
\IEEEauthorblockA{Faculty of Statistic and Computer Engineering,\\ Universidad Nacional del Altiplano de Puno, P.O. Box 291\\ Puno - Peru.\\
Email: jordanpieroborda@gmail.com }
\and
\IEEEauthorblockN{Torres-Cruz Fred}
\IEEEauthorblockA{Faculty of Statistic and Computer Engineering,\\ Universidad Nacional del Altiplano de Puno, P.O. Box 291\\ Puno - Peru.\\
Email: ftorres@unap.edu.pe}}

\maketitle

\begin{abstract}
Quantum Machine Learning (QML) shows how it maintains certain significant advantages over machine learning methods. It now shows that hybrid quantum methods have great scope for deployment and optimisation, and hold promise for future industries. As a weakness, quantum computing does not have enough qubits to justify its potential. This topic of study gives us encouraging results in the improvement of quantum coding, being the data preprocessing an important point in this research we employ two dimensionality reduction techniques LDA and PCA applying them in a hybrid way Quantum Support Vector Classifier (QSVC) and Variational Quantum Classifier (VQC) in the classification of Diabetes.

\end{abstract}


\IEEEpeerreviewmaketitle

\section{Introduction}
Machine learning is the best way to solve problems with well-defined outcomes. Image recognition, finding patterns in missing data and understanding clear and unambiguous language are all AI can do\cite{Ref1}. It is also commonly used to find differences in financial transactions, make predictions based on patterns in past data (such as the stock market) and determine when someone has sent spam and mark it as such\cite{Ref2}.

Under this premise and with the advance of quantum computers, a new field of study and research is beginning to emerge called Quantum Machine Learning (QML). The goal of quantum technologies is to demonstrate their potential in comparison to classical machine learning, but this in turn shows weaknesses such as the limitation of qubits and continuous operations of logic gates \cite{Ref3}.

While some companies have explored the use of quantum machine learning in their research and research projects, it is still quite rare to find companies that are using quantum machine learning in their business operations on a widespread basis. Some companies that have mentioned the use of quantum machine learning or have conducted research in this field include:

Google: The technology company has conducted research into the use of quantum machine learning in tasks such as route optimisation and financial data analysis.

IBM: The technology company has developed a quantum machine learning platform called IBM Q, which is used to investigate how quantum machine learning can improve the performance of tasks such as image classification and natural language processing.

Microsoft: The technology company has conducted research into using quantum machine learning to improve energy efficiency in data centres and to optimise power distribution in power grids.

D-Wave: This quantum technology company has developed a quantum machine learning platform called D-Wave Leap, which is used to investigate how quantum machine learning can improve decision-making in various industries.

In this paper, we will apply dimensionality reduction to the structure of the data set, as well as show that Linear Discriminant Analysis (LDA)\cite{Ref4}. Shows a significant advantage in supervised preprocessing over Principal Component Analysis (PCA)\cite{Ref5}.

We aim to compare Classical and Quantum classification methods - Quantum Support Vector Classifier (QSVC)\cite{Ref6} and Variational Quantum Classifier (VQC). Qiskit Machine Learning was used for the construction of fundamental computational building blocks, such as Quantum Kernels and Quantum Neural Network, which we use for the classification of Diabetes\cite{Ref7}.

\section{DATASETS}
The purpose of the data selection in this study is to categorize patients with diabetes. We utilize a CSV file to store the retrieved Kaggle dataset.
\hfill

\begin{tabular}{| c | c | c | c | }
\hline
Var & Type & Scale & Description \\ \hline
1 & V.Input & Quant. Discrete & Pregnancies \\
2 & V.Input & Quant. Continues  & Glucose \\
3 & V.Input & Quant. Continues  & P. Blood \\
4 & V.Input & Quant. Continues   & Skin-Thickness \\
5 & V.Input & Quant. Continues  & Insulin \\
6 & V.Input & Quant. Continues  & BMI \\
7 & V.Input & Quant. Continues & P.Diabetes\\
8 & V.Input & Quant. Discrete & Age \\
9 & V.Output & Quant.Dichotomous & A. "Yes" or "No \\
\hline
\end{tabular}
\hfill

Table 1 Description of variables Pima 
\hfill

Pima Indians Diabetes Database 
\hfill

\url{https://www.kaggle.com/datasets/nancyalaswad90/review}
\hfill

The aim of the data set is to identify the presence or absence of diabetes in a patient based on particular diagnostic parameters provided in the data set. There were a number of limitations used to choose these instances from a broader data set. A number of limitations were implemented as a consequence of which a set of patients who matched the description of being Pima women of at least 21 years of age were chosen as patients.

\section{METHODS APPLIED}
\subsection{Dimensionality reduction}
Dimension reduction, also known as principal component analysis (PCA), is a data processing technique used to reduce the dimensionality of a data set. PCA is based on the idea of finding a set of principal components that explain most of the variability in the data.\cite{Ref8}.

The goal is to transform a large dataset (a large number of features) into a compact representation containing important information about the data (orthogonal projection). Size reduction implies a loss of accuracy, but PCA, uses an eigenvalue suppression process to transform the covariance matrix involved in this process. Components are linear combinations of different objects to create new unallocated objects. The first element will contain the most information, then the rest in the second, and so on. The geometrical representation of the PCA are the components representing the direction of maximum change (rotation)\cite{Ref9}.

Linear discriminant analysis (LDA) is similar to PCA, these are linear transformations to reduce the dimensionality of the data set (eigenvalue separation). Where PCA maximises the variance, LDA maximises the class separation axes. LDA will create a k-dimensional subspace from the n-dimensional space of the original data, where k<=n-1. The subspace is calculated taking into account the labels to maximise class segregation.

\subsection{METRICS}
Fundamentals for evaluating various classifiers and their corresponding expressions. Where FN, TN, FP and TP are False positive, true negative, false positive and true positive.

\hfill

\vspace*{3mm}

\begin{tabular}{| c | c |}
\hline
Metrics & Equation \\ \hline
Precision & $\frac{TP}{TP+FP}$ \\
Recall & $\frac{TP}{TP+FN}$ \\
F1-Score & $\frac{2x(Precision x Recall)}{Precision + Recall}$ \\
Balanced Accuracy &($\frac{TP}{TP+FN}$$/$$\frac{TN}{TN+FP})$$/2$ \\
\hline
\end{tabular}

\vspace*{2mm}
Tabla 2.
\hfill

\subsection{Backends}
Making a choice between backends and quantum computer simulations is difficult when quick iteration over huge data sets is required.
Before entering into production or actual production, machine learning models typically need a number of modifications and iterations (remediation)\cite{Ref10}. The difficulties are the same for QML, but the hardware ecosystem is distinct. On both real devices and simulators (noisy and crowded quantum computer simulations), quantum algorithms can be used\cite{Ref11}.

\subsubsection{Simulators}

In this research, only the Qiskit Aer simulator and the standard Pennylane qubit simulator were used.

Qiskit Aer is a quantum machine simulator developed by IBM that is used to simulate the behaviour of a quantum machine on a classical computer\cite{Ref12}. Qiskit Aer is an open source tool and can be used to research and develop quantum algorithms and programs without the need to access a real quantum machine.\cite{Ref13}

PennyLane is an open source framework for quantum computation and quantum machine learning. PennyLane can be used to simulate and run quantum algorithms and programs on different quantum backends\cite{Ref14}, including quantum machine simulators such as Qiskit Aer and real quantum machines\cite{Ref15}.

\section{ALGORITHMS}
\subsection{Machine Learning Models}
Since supervised learning involves classification difficulties, we use many of the classical methods in this field of machine learning to compare the combined quantum classical approach.

\subsubsection{Regresión logística}
This approach to the binary classification issue is among the simplest. The model is trained to learn the parameters of the linear equations in their entirety:
\hfill

\begin{equation}
    \hat{k}^i = \beta_0 +\beta_1 x_1^i + \beta_2 x_2^i + \beta_n x_n^i
\end{equation} where $  \beta_n $  are the linear regression coefficients $ x_n $, - false positives, respectively, the sample characteristics. Since linear regression could not successfully achieve the classification objective, the regression formula was introduced into the logistic function as Equation (1). To determine the probability

\begin{equation}
    P(y^i = 1) = \frac{1}{1+e^-(\beta_0 +\beta_1 x_1^i + \beta_2 x_2^i + \beta_n x_n^i)}
\end{equation}

P is the probability that the label $y$ for sample $i$ is associated with the number 1. The probability threshold is set to 0.5 to change the binary output after calculating the probability of each sample (search model). ). If $P(y^i = 1) < 0.5$, the appropriate label is 0, if $P(y^i = 1) \geq 0.5$1 is the label. A sufficient number of stable samples is required for the logistic regression method to successfully estimate the parameters. $\beta_n$

\subsubsection{Classification and Regression Trees”}

A decision tree, also known as a classification and regression tree (CART), is a binary graph model in which the choice you make for the previous child determines the decision you make for the next child. The root of the tree serves as the starting point from which two branches emerge, divided into "yes" and "no" categories. Until a final choice, called a leaf, is reached, the tree structure is built up through a series of decisions. Although this approach is simple, it is prone to overfitting. These are powerful algorithms that can be successfully adapted to complex data sets. The learning process uses entropy or the Gini criterion (equation (2)).

\begin{equation}
    H_i = \sum_{k=1}^{10}P_(i,k) log_2 P_(i,k)
\end{equation}
where $i$ is the $i^{th}$ node, $P_(i,k)$, is the probability of the $k$ category..

\subsubsection{Naive Bayes}
The Nave Bayes algorithm, sometimes known as NB, is a simpler version of the Bayes theorem (Eq. 3)

\begin{equation}
    P(A|B) = \frac{P(A|B).P(A)}{P(B)}
\end{equation}

\subsubsection{k-Nearest Neighbors}
The simplest algorithm based on distance without parameters is called k-nearest neighbour (k-NN). Assume that in n-dimensional space the corresponding point will be the density. Using a distance calculation such as Euclidean distance, the point will be encoded and placed. Then, the algorithm then determines the class to apply to the new point by looking at the closest K points, averaging the classes, and predicting the appropriate class for the next new point\cite{Ref16}.

Where A and B are events, $P(A|B)$ represents the probability that A occurs if B is true, $P(B|A)$ represents the probability that B occurs if A is true, and $P ( A )$ and $P(B)$ are independent probabilities of events A and B respectively. The probabilities are conditionally independent of the classifier DS. He greatly simplified the mathematics and made it an easy problem to solve.

\subsection{Quantum Machine Learning Models}
Quantum machine learning is a subfield of machine learning that uses quantum computers to perform machine learning tasks. Quantum machine learning algorithms can be used to solve problems that are difficult or impossible to solve using classical machine learning algorithms. These problems include finding patterns in large datasets, optimizing complex functions, and classifying data.

There are several approaches to quantum machine learning, including:

Quantum support vector machines: These are quantum versions of support vector machines, which are a type of linear classifier.

Quantum principal component analysis: This is a quantum version of the principal component analysis algorithm, which is used to reduce the dimensionality of data.

\subsubsection{Quantum Kernel}
A quantum kernel is a function that is used to define the similarity between two quantum states in a quantum machine learning algorithm. The quantum kernel is a generalization of the classical kernel function, which is commonly used in kernel methods, a type of machine learning algorithm that operates by mapping data into a high-dimensional feature space.

In quantum machine learning, the quantum kernel is used to define the similarity between two quantum states in the feature space. This allows the quantum machine learning algorithm to compare and classify quantum states based on their similarity.

\subsubsection{Quantum Encoding}
The process leading from conventional data to canonical representation is known as canonical codification. There are several ways to process basic data and provide useful representations. In this study, we employ the cunning characteristic map (Qiskit ZZFeatureMap) and the angular codification (Pennylane) QSVC and VQC, respectively.

\subsubsection{Angle Encoding}
The traditional process of encoding information by rotation is called angular encoding. Traditional information is represented by rotating arrows at the appropriate ports and can be written as an equation:
\begin{center}
   {\includegraphics[width=0.25\textwidth]{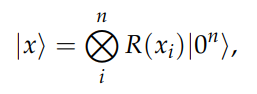}\par} 
\end{center}
where R is a rotating door such as Rx, Ry, and Rz. Angular encoding is used when the width of the eigenvector x is equal to the number of qubits.
\subsubsection{WorkFlow}
We present here the workflow used in this study to compare the selected algorithms (classical and computational). A series of algorithms are applied to the data. A rendering created using dimensionality reduction techniques. The workflow consists of four steps:
\hfill

1 Apply an Exploratory Data Analysis after gathering the data. To prepare the data for the dimensionality reduction method, the data must be cleaned and normalized with a good format.
\hfill

2 Dimensionality reduction: LDA and PCA are used to reduce the number of compressed two-dimensional functions. The PCA method uses two components. In another dataset, LDA is used. All media parts are scaled down using LDA components.
\hfill

3 Quantum encoding: Using maps of cuanic characteristics, the traditional data are codified into a cuanic representation. Only computational algorithms used this step.
\hfill

4 Models used: A combination of selected algorithms (ML and QML) is applied to the data encoding (quantum or classical) and evaluated using the same metric (see Table 1 for details).
\section{RESULTS}
The results of using the traditional classification models, such as LR, CART, KNN, and NB, are shown in this section along with a comparison to the results obtained using the machine learning statistical models, QVC and QSVC, which were used to classify diabetes.

Tab. 3: Classical models (LR, KNN, CART, NB), as well as computational models (QSVC, VQC), applied to the entire set of diabetes data using PCA dimensionality reduction.
\begin{center}
        {\includegraphics[width=0.5\textwidth]{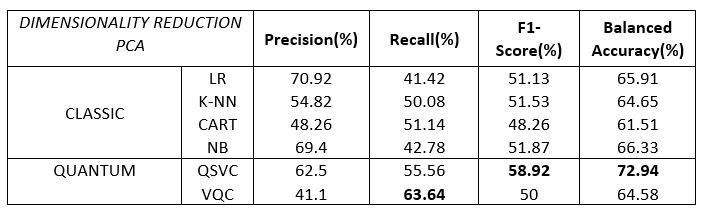}\par}
\end{center}
Tab. 4: Classical models (LR, KNN, CART, NB), as well as computational models (QSVC, VQC), applied to the entire set of diabetes data using LDA dimensionality reduction.
\vspace{0.1cm}
\begin{center}
    {\includegraphics[width=0.5\textwidth]{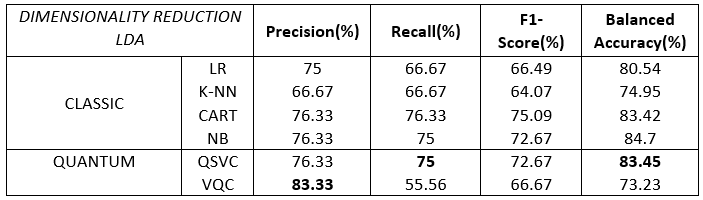}\par}
\end{center}
\vspace{0.1cm}
Figure 1. Comparison of QSVC and VQC  measures for diabetes classification using PCA and LDA.
{\includegraphics[width=0.45\textwidth]{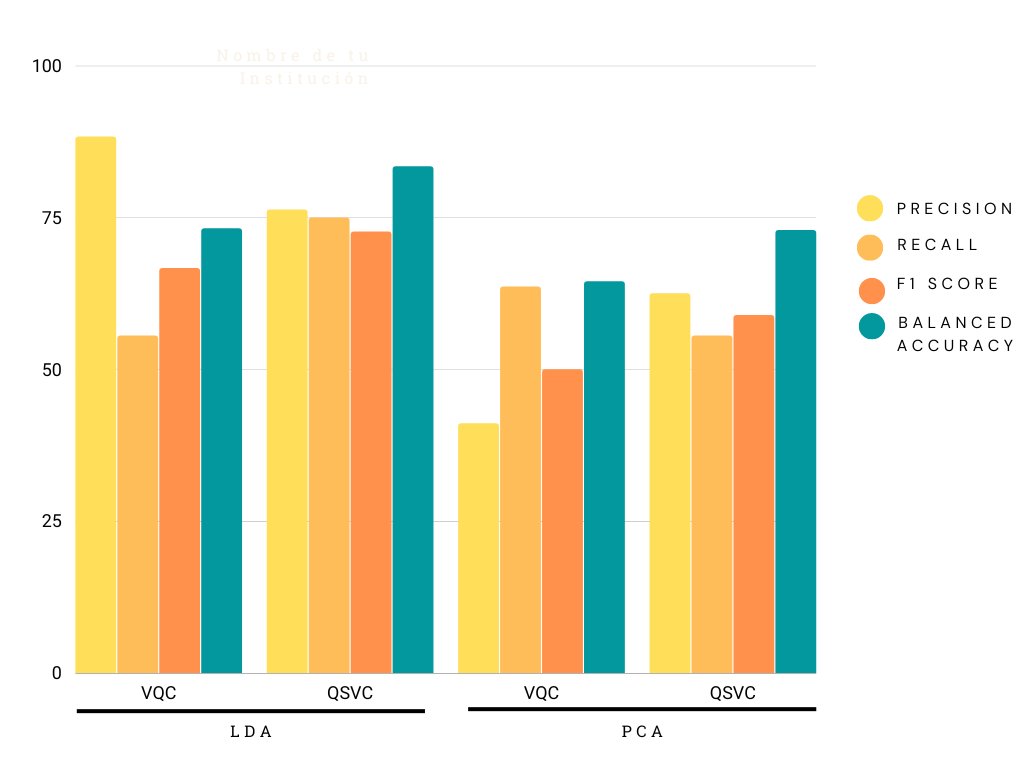}\par}
Figure 2 compares the metrics of QSVC and VQC with CART and KNN (the best traditional algorithms) when LDA is used to classify diabetes.
{\includegraphics[width=0.45\textwidth]{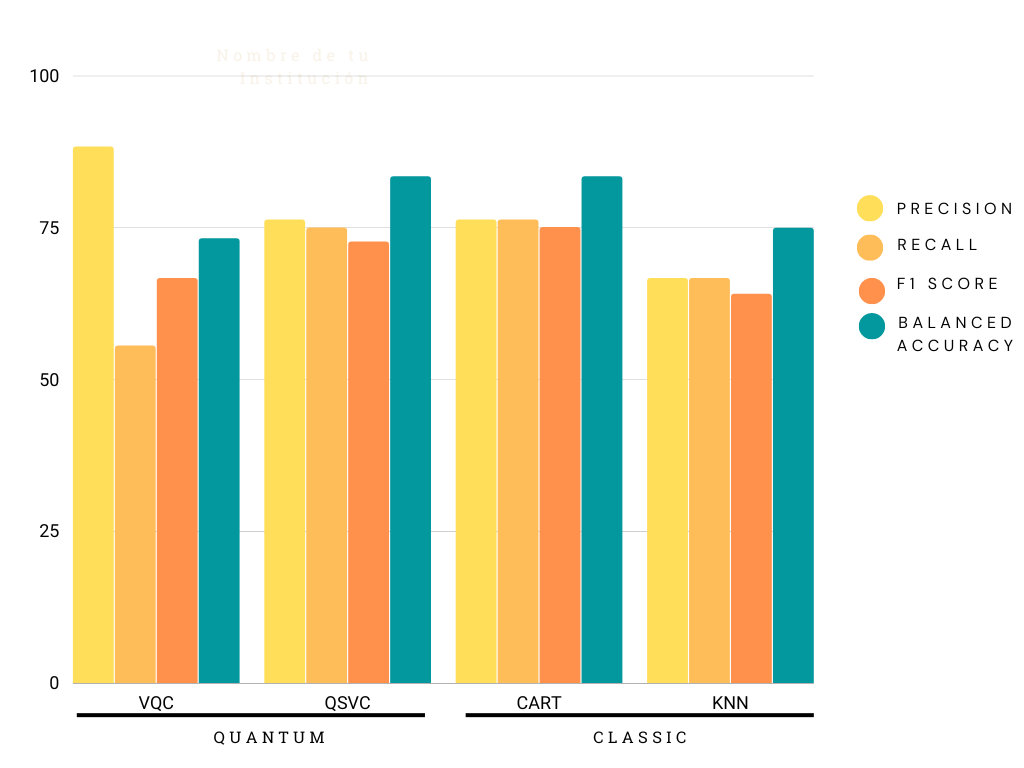}\par}
\section{DISCUSSION AND CONCLUSIONS}
We show how a quantum computer can use classification results obtained with only a subset of measurements to extract more useful information from classical data. Use in. Competitive advantage is not necessarily measured by its ability to outperform standard machine learning models, but it can be considered the best method of information extraction.
Some of the potential benefits of quantum machine learning include:

Faster speed: Quantum machine learning can perform computations much faster than classical machine learning due to the higher speed of quantum computers. For example, one study has shown that a quantum machine learning algorithm can classify data faster than a classical machine learning algorithm\cite{Ref17}.

Higher accuracy: Quantum machine learning can produce more accurate results than classical machine learning due to the higher accuracy of quantum computers. For example, one study has shown that a quantum machine learning algorithm can classify data more accurately than a classical machine learning algorithm\cite{Ref18}.

Higher learning capacity: Quantum machine learning can learn faster and more accurately than classical machine learning due to the higher processing capacity and accuracy of quantum computers. For example, one study has shown that a quantum machine learning algorithm can learn faster and more accurately than a classical machine learning algorithm\cite{Ref18}.

Increased processing power: Quantum machine learning can process a larger amount of data at the same time due to the increased processing power of quantum computers. For example, one study has shown that a quantum machine learning algorithm can process large amounts of data faster than a classical machine learning algorithm \cite{Ref19}.

For computer supervised machine learning tasks, LDA shows most promising results. To understand how LDA provides a better data representation for qubit encoding, the prevalence of LDA in PCA has not been explored in this paper, but will be considered in the future.

comparisons between LDA and PCA:

Approach: LDA focuses on maximising the separation between different classes of data while PCA focuses on maximising the variance of the data. According to one study, "LDA is a supervised method that is used to reduce the dimension of the data while maintaining as much class information as possible. On the other hand, PCA is an unsupervised method that is used to reduce the dimension of the data while maintaining as much variance as possible"\cite{Ref20}.

Number of components: LDA produces a number of components equal to the number of classes minus one, while PCA produces a number of components equal to the number of variables minus one. According to one study, "LDA always produces the number of components equal to the number of classes minus one, while PCA produces the number of components equal to the number of variables minus one. Therefore, LDA is more suitable for the analysis of data with a small number of classes, while PCA is more suitable for the analysis of data with a large number of variables"\cite{Ref20}.

Data requirements: LDA requires data to be normally distributed and classes to have equal covariance matrices, whereas PCA does not have these requirements. According to one study, "LDA requires data to be normally distributed and classes to have equal covariance matrices. On the other hand, PCA does not have these requirements. Therefore, LDA is more sensitive to the violation of these assumptions than PCA" \cite{Ref20}.

\end{document}